\documentclass[conference]{IEEEtran}
\IEEEoverridecommandlockouts
\usepackage{cite}
\usepackage{amsmath,amssymb,amsfonts}
\usepackage{algorithmic}
\usepackage{graphicx}
\usepackage{textcomp}
\usepackage{xcolor}
\usepackage{booktabs}
\usepackage[]{multirow}
\usepackage{enumitem}
\usepackage[autostyle]{csquotes}
\usepackage{tabulary}
\usepackage{ragged2e}
\usepackage{longtable,array,ragged2e}
\newcolumntype{L}[1]{>{\raggedright\let\newline\\\arraybackslash\hspace{0pt}}m{#1}}
\newcolumntype{C}[1]{>{\centering\let\newline\\\arraybackslash\hspace{0pt}}m{#1}}
\newcolumntype{R}[1]{>{\raggedleft\let\newline\\\arraybackslash\hspace{0pt}}m{#1}}
\usepackage{graphicx}
\usepackage{booktabs,dcolumn,caption}
\newcolumntype{d}[1]{D{.}{.}{#1}} 
\renewcommand{\ast}{{}^{\textstyle *}} 
\usepackage{bookmark}%
\def\BibTeX{{\rm B\kern-.05em{\sc i\kern-.025em b}\kern-.08em
    T\kern-.1667em\lower.7ex\hbox{E}\kern-.125emX}}
\begin{document}

\title{Natural Language Generation Using Reinforcement Learning with External Rewards\\
}

\author{\IEEEauthorblockN{Vidhushini Srinivasan\IEEEauthorrefmark{1}, Sashank Santhanam\IEEEauthorrefmark{2}, Samira Shaikh\IEEEauthorrefmark{4}}
\IEEEauthorblockA{\textit{Department of Computer Science} \\
\textit{University of North Carolina at Charlotte}\\
Charlotte, USA \\
\{\IEEEauthorrefmark{1}vsriniv6, \IEEEauthorrefmark{2}ssantha1, \IEEEauthorrefmark{4}samirashaikh\}@uncc.edu
}
\and
}

\maketitle

\begin{abstract}
We propose an approach towards natural language generation using bidirectional encoder-decoder which incorporates external rewards through reinforcement learning (RL). We use attention mechanism and maximum mutual information as initial objective function using RL.  
Using a two-part training scheme, we train an external reward analyzer to predict the external rewards and then use the predicted rewards to maximize the expected rewards (both internal and external). We evaluate the system on two standard dialogue corpora - Cornell Movie Dialog Corpus and Yelp Restaurant Review Corpus. We report standard evaluation metrics including BLEU, ROUGE-L and perplexity as well as human evaluation to validate our approach. \end{abstract}


\begin{table*}[!ht]
\small
\centering
\begin{tabular}{@{}llll@{}}
\toprule
\multicolumn{1}{c}{\textbf{Input Text}} & \multicolumn{1}{c}{\textbf{Human Ground Truth}} & \multicolumn{1}{c}{\textbf{\begin{tabular}[c]{@{}c@{}}Generated \\ response \\ without RL\end{tabular}}} & \multicolumn{1}{c}{\textbf{Generated response with RL}} \\ \midrule
\begin{tabular}[c]{@{}l@{}}Very friendly staff and a great range. \\ Their prices may be few dollars more \\ than a box store, but I like to support \\ a local business and they definitely \\ earn it in service.\end{tabular} & \begin{tabular}[c]{@{}l@{}}As long as you are safe \\ (and they do watch you) they \\ don't bother you. I love that \\ and wish more ranges did that. \\ Go, be safe, have fun,\\  and enjoy your rights.\end{tabular} & I don't understand. & \begin{tabular}[c]{@{}l@{}}The limited menu helps them keep their \\ focus on the freshness of the meal. \\ This place is always busy and is a \\ testament to their product.  \\ Go. Eat. Enjoy. This place is awesome!!\end{tabular} \\ \midrule
\begin{tabular}[c]{@{}l@{}}does not sound free to me.\\ so you want me to pay for it ?\end{tabular} & Yes, I think so. & I don't know. & Yes sir! I`m sure you do have to rush. \\ \bottomrule
\end{tabular}
\caption{\label{tab:response-generation}Example generated text from both the corpora  (Yelp Review dataset in first row and Cornell Movie Dialog in second row). Our proposed system with Reinforcement Learning (RL) is able to match the emotional tone of the context provided in the input text.}
\end{table*}
\section{Introduction}
We aim to develop models that are capable of generating language across multiple genres of text \textendash~say, conversational text and restaurant reviews. After all, humans are adept at both. Extant natural language generation (NLG) models work on either conversational text (e.g. movie dialogues) or longer text (e.g. stories, reviews) but not both \cite{guo2017long}, \cite{D16-1127}. In addition, while the state-of-the-art in this field has advanced quite rapidly, current models are prone to generate language that is short, dull, off-context or vague. More importantly, the generated language may not adequately reflect the affective content of the input. Indeed, humans are adept at this task as well. 
To address these research challenges, we propose an RNN-LSTM architecture that uses an encoder-decoder network. We also use reinforcement learning that incorporates internal and external rewards. Specifically, we use emotional appropriateness as an internal reward for the NLG system \textendash~so that the emotional tone of the generated language is consistent with the emotional tone of prior context fed as input to the model.  We also effectively incorporate usefulness scores as external rewards in our model. Our main contribution is the use of distantly labeled data in an architecture that generates coherent, affective content and we test the architecture across two different genres of text.

\section{Problem Statement and Intuition}
Our goal is to take advantage of reinforcement learning and external rewards during the process of language generation. Complementary to this goal, we also aim to generate language that has the same emotional tone as the preceding input. Emotions are recognized as functional in decision-making by influencing motivation and action selection \cite{Moerland2018}. 
However, external feedback and rewards are hard to come by for language generation; these would need to be provided through crowdsourcing judgments on the generated responses \textit{during} the generation process, which makes the process time-consuming and impractical. To overcome this issue, we look for distance labeling \cite{mintz2009distant}  - and use labels provided in the training set as a proxy for human judgments on the generated responses. Specifically, we incorporate usefulness scores in a restaurant review corpus as a proxy for external feedback. 

\begin{figure}[htbp]
\centering
    \includegraphics[width=3.05in, height=2in]{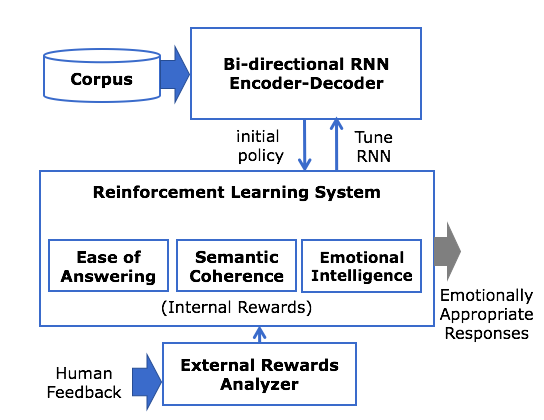}
\caption[Overall Architecture]{\label{fig:dist}Overall Architecture of the system showing internal and external rewards using reinforcement learing}
\end{figure}

\section{Model Architecture}
Figure~\ref{fig:dist} shows overall system architecture. We  use a bidirectional RNN \textit{seq2seq} encoder-decoder network with maximum mutual information as objective function. We tune the system using reinforcement learning with internal reward functions to promote ease of answering, semantic coherence \cite{D16-1127} along with emotional intelligence \cite{asghar2018affective} and use external rewards from human feedback to generate emotionally rich responses. Table 1 shows examples of system output on two different corpora, the Cornell Movie Dialog Corpus and the Yelp review dataset.

\subsection{Affective Word Embeddings}
We use the Affective Norms of Words (ANEW) lexicon \cite{Warriner2013} that has valence, arousal and dominance scores for words to augment existing word embeddings \cite{NIPS2013_5021}. We append Valence (V), Arousal (A) and Dominance (D)  score from the ANEW lexicon to each word, resulting in 1027 dimensions for each word. 
In cases where a match cannot be found in the lexicon, we append a neutral vector [5,1,5] similar to \cite{asghar2018affective}. This word2vec-VAD embedding is fed as input to the bidirectional RNN encoder-decoder seq2seq model and is also used in the RL system to model the Emotional Intelligence heuristic.

\subsection{Adaptive RL System}

\begin{table*}[!ht]
\small
\centering
\begin{tabular}{@{}llll@{}}
\toprule
 & \multicolumn{1}{c}{\textbf{Objective Function}} & \multicolumn{1}{c}{\textbf{\begin{tabular}[c]{@{}c@{}}Internal Rewards\end{tabular}}} & \multicolumn{1}{c}{\textbf{External Rewards}} \\ \midrule
\begin{tabular}[c]{@{}l@{}}\textbf{Li et al. Approach}\end{tabular} & \begin{tabular}[c]{@{}l@{}}Policy Gradient Method\end{tabular} & \parbox{4cm}{\begin{itemize}
    \item Ease of Answering
    \item Information Flow
    \item Semantic Coherance
    \end{itemize} }
& \begin{tabular}[c]{@{}l@{}} N/A \end{tabular} \\ \midrule
\begin{tabular}[c]{@{}l@{}}\textbf{Our Proposed Approach} \end{tabular} & \textbf{Maximum Mutual Information (MMI)} & 
\parbox{4cm}{ \begin{itemize}
    \item Ease of Answering
    \item Semantic Coherance
    \item \textbf{Emotional Intelligence} \end{itemize} }
  & \begin{tabular}[c]{@{}l@{}} \textbf{Human Feedback}   \end{tabular}\\ \bottomrule
\end{tabular}
\caption{\label{tab:response-generation}State-of-the-art Method vs. Our Proposed Approach. We use a different Objective Function, and Internal as well as External Rewards in our model}
\end{table*}

We use a bidirectional RNN encoder-decoder seq2seq model \cite{vinyals2015neural} with Bahdanau-style attention mechanism \cite{bahdanau2014neural}. 
Next, we use a greedy decoder to generate the best response at every stage of decoding during decoder training and inference phases. We fine tune the basic \textit{seq2seq} generative model with the internal and external rewards in our RL tuner system to generate more interesting, diverse and emotionally appropriate responses. The internal rewards take into account coherence, flexibility in answering and also emotional intelligence measures whereas external rewards incorporates human feedback to make the responses resemble human produced ones. Our approach is very closely related to Li \emph{et al.}  \cite{D16-1127}, however, with key differences in the objective functions and the use of external rewards. These are highlighted in Table~\ref{tab:response-generation}. 

The standard objective function for seq2seq models is the log-likelihood of target T given source S, given as follows:

\setlength{\abovedisplayskip}{1pt}
\setlength{\belowdisplayskip}{2pt}
\begin{equation}
 \hat{T} = \arg\max_T \{\log p(T | S) \}
\end{equation}
This formulation leads to generic responses, since it only selects for target given source. We optimize this standard objective function by replacing it with Maximum Mutual Information (MMI) \cite{N16-1014}. In MMI, parameters are chosen to maximize (pairwise) mutual information
between the source S and the target T:

\setlength{\abovedisplayskip}{3pt}
\setlength{\belowdisplayskip}{3pt}
\begin{equation}
  \dfrac{\log p(S, T)}{p(S)p(T)} 
\end{equation}

Doing so avoids favoring responses that unconditionally enjoy high probability, and instead biases towards those responses that are specific to the given input. The MMI objective can written as follows:

\begin{equation}
\hat{T}=\arg\max_T \{\log p(T|S)-\lambda \log p(T)\}
\end{equation}

Here,  $\lambda$ is the hyperparameter that controls the extent to which we penalize generic responses to get more diverse responses.
Adjusting the value of $\lambda$ results in a reasonable number of diverse responses, however, these could still be dull and also lack emotion and proper grammatical structure. 
To address these issues, we model the reinforcement learning system with appropriate heuristics. 

The generated sentences from the \textit{seq2seq} model can be viewed as actions that are taken by the policy defined by the encoder-decoder language model. The parameters of this encoder-decoder network are fine-tuned using reinforcement learning with policy gradient method \cite{D16-1127}. The components are:

\textit{Action (a)}~\textendash~dialog utterances to generate i.e. action $a = gen(S)$, where $gen(S)$ is the sequence generated by RNN-LSTM. The action space is infinite and generates sequences of varying length.

\textit{State (S)}~\textendash~dialog is transformed to a vector representation by feeding the current dialog (state) for which the response has to be generated.

\textit{Policy}~\textendash~policy takes the form $p_{RL}(p_{i+1}|p_{i})$ where  $p_{i+1}$ is the response to be generated for the given dialog $p_{i}$. Here, we use a stochastic distribution to represent policy as it is the probability distribution over actions given states, where both state and actions are dialogs. By doing so, we overcome the difficulty of optimizing a deterministic policy, as that would lead to discontinuous objective and cannot be further used with gradient-based methods.

\begin{figure}[htbp]
 \begin{centering}
\includegraphics[height=2.5in, width=3in]{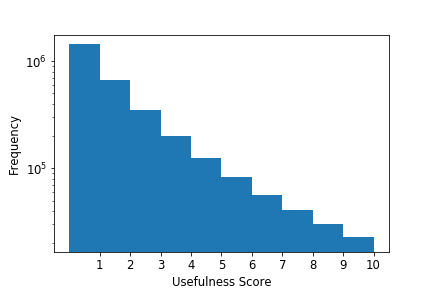}
 \par\end{centering}

 \caption[Yelp-Useful score distribution]{\label{fig:yelp-dist}Yelp-Useful score distribution.}
 \end{figure}

\textit{Rewards (r)}~\textendash~We implement three internal rewards and one external reward to overcome the issues in generating language with seq2seq architecture \cite{vinyals2015neural}. The three internal are Ease of Answering $r_{EA}$, Semantic Coherence $r_{SC}$, Emotional Intelligence $r_{EI}$ \cite{D16-1127} and one external reward \cite{NIPS2017_7017} from human feedback  $r_{HF}$.
\begin{enumerate}[nolistsep,noitemsep]
    \item \textit{Ease of Answering (EA)} ($r_{EA}$) \textendash~is measured as negative log likelihood of generating a dull response for a dialog. Following \cite{D16-1127},  \cite{DBLP:journals/corr/LowePSP15} and \cite{niu2018polite}, we compose a list of 10 dull responses that frequently occur in the seq2seq model and penalize the model when it generates those responses.\footnote{The 10 responses are: \textit{``I don't know.", ``I don't know what I mean.", ``I don't know what you're talking about.", ``You don't know.", ``You know what I mean.", ``You know what I'm saying.", ``You don't know anything.", ``I am not sure.", ``I know what you mean.", ``I do not know anything."}}
     Let set $S$ represent a list of dull responses. Then, the reward function can be defined as follows:
     
     \begin{equation}
      r_{EA} = -\dfrac{1}{N_{S}}\Sigma\dfrac{1}{N_{S}}\log p_{seq2seq(s|a)}
      \end{equation}
      
      $p_{seq2seq}$ represents likelihood output of \textit{seq2seq} model. The RL system is likely to penalize utterances in the above composed list and hence less likely to generate dull responses. $r_{EA}$ is scaled by length of the target S.\\
     \item \textit{Semantic Coherence (SC)} ($r_{SC}$) \cite{D16-1127} \textendash~is used to avoid situations in which the generated responses are highly rewarded, but are neither grammatical nor coherent. We consider the mutual information between the action $a$ and the given input to ensure that the responses are coherent and appropriate. This also involves reverse training the model where we count the probability of the input prompt given the current generated response. 

     \begin{equation}
     \begin{array}{l}
     r_{SC} = \dfrac{1}{N_{y}}\log p_{seq2seq}(y|x_{i})+\\
     \dfrac{1}{N_{x_{i}}}\log p_{backward-seq2seq}(x_{i}|y)
     \end{array}
     \end{equation}
     \item \textit{Emotional Intelligence (EI)} ($r_{EI}$) \cite{asghar2018affective} \textendash~This reward is incorporated by minimizing affective dissonance between the prompts and the responses. This approach tries to maintain affective consistency between input and generated response. The heuristic is based on the fact that open-domain textual conversations between humans follow an affective pattern. Thus, we make an assumption that the affective tone does not fluctuate often in general and we focus on minimizing the dissonance in affective tone between the input prompt and the generated responses.
     
     \begin{equation}\label{eq:w2av}
     \resizebox{.91 \hsize}{!}{%
     $ 
     \begin{array}{l}
     r_{EI_{i}} = \lambda p(a)\lVert {\sum_{j=1}^n\dfrac{W2AV(x_{j})}{|X|} - 
     \sum_{k=1}^i\dfrac{W2AV(y_{k})}{i}}\rVert
     \end{array}
     $
     }
     \end{equation}
     
      Here, \textit{W2AV} in Equation \ref{eq:w2av} denotes the word-affect vector of the given sequence and the term $\sum_{j=1}^n \dfrac{W2AV(x_{j})}{|X|}$ denotes average affect vector of the input prompt and $\sum_{k=1}^i \dfrac{W2AV(y_{k})}{i}$ denotes average affect vector of the generated response up to the current step $i$.\\
    
\item \textit{Human Feedback (HF)} ($r_{HF}$) \textendash~To incorporate external rewards in our model, the external rewards analyzer is trained with human feedback. We simulate human feedback through the reviews from the Yelp dataset usefulness score. We categorize each review in the Yelp dataset into two main classes {\fontfamily{qcr}\selectfont
Useful} and {\fontfamily{qcr}\selectfont
Not Useful}  based on the frequency distribution of the reviews (as shown in Figure~\ref{fig:yelp-dist}). Reviews with normalized scores $<5$ are considered not useful, while the rest are considered to be useful. We exclude all reviews that do not have usefulness ratings, since it is not clear which category they would fall under. 
Next, we train an SVM classifier to differentiate between the two classes {\fontfamily{qcr}\selectfont
Useful}  and {\fontfamily{qcr}\selectfont
Not Useful} as described above. During the training phase, we determine whether the generated response is useful or not (by classifying the generated output in real-time using the SVM classifier) and give the reward accordingly. This synthetic feedback \cite{NIPS2017_7017} from the external reward analyzer is provided throughout the training phase and a greedy decoder is then used to generate the best response.
\end{enumerate}

\section{Experiments and Results}
We test the efficacy of the proposed method in generating text in two different corpora, which pertain to different genres of text. The corpora we used are the Cornell Movie Dialog corpus \cite{Danescu-Niculescu-Mizil+Lee:11a} and the Yelp Restaurant Review dataset.  The Cornell Movie-Dialog corpus \cite{Danescu-Niculescu-Mizil+Lee:11a} contains a large metadata-rich collection of fictional conversations extracted from raw movie scripts. There are 220,579 conversational exchanges between 10,292 pairs of movie characters involving 9,035 characters. There are 304,713 utterances in total. The Yelp review dataset contains 5.9M reviews. Along with the reviews, this  dataset contains nine additional features, including usefulness score, which we use as external rewards. We perform the standard pre-processing steps on both the Cornell and Yelp dataset, including lowercasing all conversations, expanding contractions, compress duplicate end punctuation to one symbol and removing HTML entities.  

Table ~\ref{tab:statistics} shows the descriptive statistics for both corpora. We take the
most common 12,000 words from the training and validation sets as our vocabulary as in \cite{asghar2018affective}, and replace any other tokens in these sets with an unknown symbol
$<$UNK$>$. We partition the training and validation sets such that none of the responses in the training set have $<$UNK$>$. This effectively prevents the model from generating the unknown token during inference. 
The word count is set to maximum threshold of 20.


\begin{table}[h]
\small
\centering
\begin{tabular}{@{}llll@{}}
\toprule
                                                         & \begin{tabular}[c]{@{}l@{}}Training\\ set\end{tabular} & \begin{tabular}[c]{@{}l@{}}Validation\\ set\end{tabular} & \begin{tabular}[c]{@{}l@{}}Testing \\ set\end{tabular} \\ \midrule
\begin{tabular}[c]{@{}l@{}}Cornell corpus\end{tabular} & 160,000                                                & 14,000                                                   & 6000                                                   \\ \midrule
\begin{tabular}[c]{@{}l@{}}Yelp corpus\end{tabular}    & 4,017,986                                              & 1,187,406                                                & 791,604                                                \\ \bottomrule
\end{tabular}
\caption{\label{tab:statistics}Descriptive statistics for the two corpora used in our experiments}
\end{table}


\subsection{Proposed Model \& Baseline}
We use a baseline which is a basic seq2seq model with MMI objective function \cite{N16-1014}, that does not use reinforcement learning. Our proposed model uses seq2seq to choose initial policy and fine-tunes that model to generate more diverse responses based on internal and external rewards as described in Section 2.2.

For Cornell corpus, the final reward function just uses the internal reward components during reinforcement learning and can be described as follows:

\setlength{\abovedisplayskip}{0pt}
\setlength{\belowdisplayskip}{0pt}
\begin{equation}
    r_{Final\_Cornell} = \lambda_1 r_{EA}+\lambda_2 r_{SC}+\lambda_3 r_{EI}
\end{equation}

For the Yelp corpus, the final reward function, is the weighted sum of both internal and external rewards, as follows:

  \begin{equation}
  r_{Final\_Yelp} = \lambda_1 r_{EA}+\lambda_2 r_{SC}+
  \lambda_3 r_{EI}+\lambda_4 r_{HF}
  \end{equation}
~\\
For both models, the values of the hyperparameters are given in Table~\ref{hyperparameters}. We can see certain differences in the hyperparameters since the Yelp corpus size is greater than the Cornell corpus (e.g. batch size and number of epochs). The learning rate and decay rate are greater for Yelp because it takes a longer time to converge and train the model than it does on the smaller corpus Cornell. The values for the rewards are adjusted and fine-tuned based on the outcome of each model.

\begin{table}[th]
\small
\centering
\begin{tabular}{@{}lll@{}}
\toprule
\begin{tabular}[c]{@{}l@{}}Hyper-\\ parameter\end{tabular} & \begin{tabular}[c]{@{}l@{}}Cornell\\ model value\end{tabular} & \begin{tabular}[c]{@{}l@{}}Yelp \\ model value\end{tabular} \\ \midrule
Batch size                                                 & 128                                                           & 512                                                         \\ \midrule
Gradient clip                                              & 1.0                                                           & 1.0                                                         \\ \midrule
Learning rate                                              & 0.01                                                          & 0.15                                                        \\ \midrule
Decay rate                                                 & 0.0095                                                        & 0.01                                                        \\ \midrule
Epochs                                                     & 50                                                            & 75                                                          \\ \midrule
LSTM layers                                                & 2                                                             & 2                                                           \\ \midrule
Encoder RNN size                                           & 1027                                                          & 1027                                                        \\ \midrule
Decoder RNN size                                           & 1027                                                          & 1027                                                        \\ \midrule
$r_{EA} \lambda_1$                                                         & 0.25                                                          & 0.25                                                        \\ \midrule
$r_{SC} \lambda_2$                                                       & 0.35                                                          & 0.25                                                        \\ \midrule
$r_{EI} \lambda_3$                                                       & 0.40                                                          & 0.25                                                        \\ \midrule
$r_{HF} \lambda_4$                                                      & \textemdash                                                            & 0.25                                                        \\ \bottomrule
\end{tabular}
\caption{\label{hyperparameters}Hyperparameter settings for the models with Reinforcement Learning used in our approach.} 
\end{table}

\begin{table}[t]
\small
\centering
\begin{tabular}{@{}lllll@{}}
\toprule
                                                                           &            & BLEU  & ROUGE-L & Perplexity \\ \midrule
\multirow{2}{*}{\begin{tabular}[c]{@{}l@{}}Cornell \\ Corpus\end{tabular}} & without RL & 0.15  & 0.39    & 98.96      \\ \cmidrule(l){2-5} 
                                                                           & with RL    & \textbf{0.38**}  &\textbf{ 0.55***}    & \textbf{76.65*}      \\ \midrule
\multirow{2}{*}{\begin{tabular}[c]{@{}l@{}}Yelp \\ Corpus\end{tabular}}    & without RL & 0.014 & 0.24    & 99.04      \\ \cmidrule(l){2-5} 
                                                                           & with RL    & \textbf{0.21**}  & \textbf{0.32**}    & \textbf{85.34**}     \\ \bottomrule
\end{tabular}
\caption{\label{tab:metric}Model evaluation on automated metrics. \\ * $p<0.05$ **$p<0.01$ ***$p<0.005$} 
\end{table}

\subsection{Performance on Automated Metrics}
We evaluate the model using automated metrics including BLEU score, ROUGE-L and Perplexity\cite{papineni2002bleu}. In Table~\ref{tab:metric}, we report scores on the automated metrics, BLEU, ROUGE-L and Perplexity. The scores are statistically significantly better than baseline (without RL), with $p<0.01$ for BLEU score, $p<0.05$ for Perplexity and $p<0.005$ for ROUGE-L for Cornell.  For the Yelp corpus, the model with external rewards performs significantly better on all three metrics ($p<0.01$) when compared to the baseline (without RL). 

\subsection{Human Evaluation of Performance}
To evaluate the performance of our model with human ratings, we performed two rounds of crowd-sourced annotation. 

First, we created a simple survey, containing 20 prompt/response pairs from both Cornell and Yelp models. We recorded a total of 52 responses from undergraduate and graduate Computer Science students. Each response generated by the system was evaluated on three measures -  \textbf{Syntactic Coherence} \textit{(how grammatical and coherent the responses are with respect to the given prompt)}, \textbf{Natural Flow} \textit{(how natural the response is to the given prompt)} and \textbf{Emotional Appropriateness} \textit{(captures the emotional  appropriateness of the text to the given prompt)} \cite{asghar2018affective}.

Next, we conducted experiments on Amazon Mechanical Turk with 100 prompts/response pairs of both Cornell and Yelp models. Each response was rated by at least 5 workers on measures of Syntactic Coherence, Natural Flow and Emotional Appropriateness. Table~\ref{tab:human_eval_metrics} shows the ratings obtained from human evaluation on the metrics of Syntactic Coherence, Natural Flow and Emotional Appropriateness. We find that our model achieves better ratings on all three metrics as we generate longer sentences for the Yelp review and the model is also able to outperform the current state of the art of the model\cite{asghar2018affective} as demonstrated in Table~\ref{tab:against_ashgar}.

\begin{table}[h]
\small
\centering
\begin{tabular}{@{}lllll@{}}

               \toprule                                                            &        & \begin{tabular}[c]{@{}l@{}}
                                                                           Syntactic \\ Coherence\end{tabular} & \begin{tabular}[c]{@{}l@{}}Natural\\ Flow\end{tabular} & \begin{tabular}[c]{@{}l@{}}Emotional\\ Appropriateness\end{tabular} \\ \midrule
\multirow{2}{*}{\begin{tabular}[c]{@{}l@{}}Cornell \\ Corpus\end{tabular}} & Survey & 1.45                                                           & \textbf{1.42}                                                   & 1.44                                                                \\ \cmidrule(l){2-5} 
                                                                           & MTurk  & \textbf{1.49}                                                           & 1.41                                                   & \textbf{1.53}                                                                \\ \midrule
                                                                           \midrule
\multirow{2}{*}{\begin{tabular}[c]{@{}l@{}}Yelp \\ Corpus\end{tabular}}    & Survey & 1.46                                                           & \textbf{1.52}                                          & \textbf{1.73}                                                       \\ \cmidrule(l){2-5} 
                                                                           & MTurk  & \textbf{1.51}                                                  & 1.50                                                   & 1.66                                                                \\
                                                                           \bottomrule
\end{tabular}
\caption{\label{tab:human_eval_metrics}Human evaluation of our models performance on measures of syntactic coherence, naturalness of flow and emotional appropriateness of generated response. Scores are averages on a 3-point (0 being lowest and 2 being highest) Likert scale, with higher scores indicating better performance on a given metric.} 
\end{table}

\begin{table}[h]
\small
\centering
\begin{tabular}{@{}llll@{}}
\toprule
                                                                      & \begin{tabular}[c]{@{}l@{}}Syntactic \\ Coherence\end{tabular} & \begin{tabular}[c]{@{}l@{}}Natural\\ Flow\end{tabular} & \begin{tabular}[c]{@{}l@{}}Emotional\\ Appropriateness\end{tabular} \\ \midrule
\begin{tabular}[c]{@{}l@{}}Cornell Model\end{tabular} & 1.49                                                           & 1.41                                                  & 1.53                                                                \\ \midrule
\begin{tabular}[c]{@{}l@{}}Ashgar \textit{ et al.}\\ (2018)\end{tabular}     & 1.45                                                           & 1.31                                                   & 1.33                                                                \\ \bottomrule
\end{tabular}
\caption{\label{tab:against_ashgar}Comparison of our model against best performing model from state-of-the-art baseline Asghar \textit{et al.} (\cite{asghar2018affective}) on measures of syntactic coherence, naturalness of flow and emotional appropriateness of generated response. Scores are averages on a 3-point (0 being lowest and 2 being highest) Likert scale, with higher scores indicating better performance on a given metric. 
}
\end{table}


From our results, we can observe that incorporating external rewards results in higher (on average) scores across our metrics of Syntactic Coherence, Natural Flow and Emotional Appropriateness than when external rewards are not incorporated in the model. 


\section{Related Work}

\textit{\textbf{Language Generation using Reinforcement Learning:}}
Our work closely follows that of Li et al. \cite{D16-1127}, who proposed an advancement in seq2seq models using reinforcement learning to obtain diverse, coherent responses that could sustain conversations. They designed appropriate reward functions to overcome some of the challenges in traditional seq2seq models. Li et al. \cite{D17-1230} also proposed using adversarial training for open-domain dialogue generation. They cast the task as a reinforcement learning problem where they jointly trained a generative model to produce response sequences, and a discriminator to distinguish between the human-generated dialogues and the machine-generated ones. Similar to our proposed method, \cite{N16-1014} used Maximum Mutual Information (MMI) as the objective function.  More recently, Sankar and Ravi \cite{sankar2019deep} propose the usage of using discrete attributes such as sentiment, dialog-acts and emotion to generate responses through the use of reinforcement that leads to improvement over traditional seq2seq models. However, in none of these prior works where any external rewards incorporated during the reinforcement learning phase. 


\textit{\textbf{Incorporating External Rewards:}} Christiano et al. \cite{NIPS2017_7017} used external rewards to fine-tune their reinforcement learning model. However, their system was trained for Atari games and simulated robot locomotion, not language generation. Perhaps the closest work to ours is the work by Niu and Bansal \cite{niu2018polite} who generated polite language by assigning rewards proportional
to the politeness classifier score of the sampled response. Their work, however,does not include emotional appropriateness. 

\textit{\textbf{Generation of Emotional Language:}} Emotions are recognized as functional in decision-making by influencing motivation and action selection. Therefore, computational emotion models are usually grounded in the agent`s decision-making architecture, of which reinforcement learning is an important subclass. Moerland \cite{Moerland2018} provides the first survey of computational models of emotion in reinforcement learning agents. The survey focuses on agent/robot emotions. Badoy and Teknomo \cite{Badoy2014QLearningWB} proposed using four basic emotions: joy, sadness, fear, and anger to influence a Qlearning agent. Simulations show that the proposed affective agent requires fewer steps to find the optimal path. 

With respect to language generation, Asghar et al. \cite{asghar2018affective} incorporated affective content in neural models by using the ANEW lexicon  \cite{Warriner2013} and appending word embeddings with affective objective functions to achieve affective response generation. Zhou et al. \cite{zhou2017emotional} have proposed Emotional Chatting Machine that can generate appropriate responses not only in content (relevant and grammatical) but also in emotionally consistent with the input prompt. However, these prior approaches also do not incorporate external feedback as a reward towards generating emotionally rich, coherent and useful language.



\section{Discussion and Future Work}

The novelty of our approach lies in the addition of Emotional Intelligence as an internal reward function and combining both internal and external rewards to create an emotionally appropriate model. The use of external rewards to generate more sensible and human-like responses is novel in natural language generation task, with the exception of recent work conducted by Niu and Bansal \cite{niu2018polite}.  

Our approach is also able to generate language across two different genres of text, one for conversational agent trained on movie dialogue and the other for Yelp restaurant reviews.

We determine through our experiments, that the our seq2seq model with MMI function and appropriately designed reward functions could generate diverse, coherent and emotionally appropriate responses.  Moreover, the metrics like BLEU, perplexity and ROUGE-L are inadequate measures of how well the model performs. Through human evaluation, we determine that model performs well and outperforms the state-of-the-art baseline in measures of syntactic coherence, naturalness of flow and emotional appropriateness. 

In future, we plan to experiment with different heuristics like maximizing affective dissonance and content as emotional intelligence heuristic reward system. We have used useful score in the Yelp restaurant review dataset as external feedback. We also plan to incorporate direct human feedback into the training phase. All the code used in these experiments and repository of additional examples is available at \textcolor{blue}{\url{https://github.com/VidhushiniSrinivasan16/ICMLA}}.

\end{document}